\documentclass{article}

\usepackage{arxiv}

\usepackage[utf8]{inputenc} 
\usepackage[T1]{fontenc}    
\usepackage{hyperref}       
\usepackage{url}            
\usepackage{booktabs}       
\usepackage{amsfonts}       
\usepackage{nicefrac}       
\usepackage{microtype}      
\usepackage{graphicx}
\usepackage[numbers]{natbib}
\usepackage{doi}

\usepackage[all]{nowidow}
\usepackage{float}
\usepackage{xcolor}
\usepackage{multirow, makecell}
\usepackage{tikz}
\usetikzlibrary{positioning,decorations.pathreplacing,calligraphy}
\usepackage{subcaption}
\usepackage{algorithm}
\usepackage{algorithmic}
\usepackage{amsmath}
\usepackage{graphicx}
\usepackage{graphbox}
\usepackage{subcaption}
\usepackage[export]{adjustbox}

\title{CNN-based explanation ensembling for dataset, representation and explanations evaluation}

\author{ \href{https://orcid.org/0000-0003-2903-6050}{\includegraphics[scale=0.06]{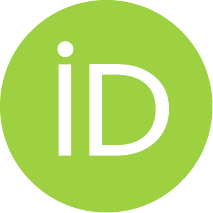}\hspace{1mm}Weronika Hryniewska-Guzik} \\
	Faculty of Mathematics and Information Science\\
	Warsaw University of Technology, Poland\\
	\texttt{weronika.hryniewska.dokt@pw.edu.pl} \\
	\And
    \href{https://orcid.org/0000-0002-2718-5426}{\includegraphics[scale=0.06]{orcid.pdf}\hspace{1mm}Luca Longo}\\
	The Artificial Intelligence and Cognitive Load Research Lab\\
	Technological University Dublin, Ireland\\
	\texttt{luca.longo@tudublin.ie} \\
	\And
	\href{https://orcid.org/0000-0001-8423-1823}{\includegraphics[scale=0.06]{orcid.pdf}\hspace{1mm}Przemysław Biecek} \\
	Faculty of Mathematics and Information Science\\
	Warsaw University of Technology\\
	\texttt{przemyslaw.biecek@pw.edu.pl} \\
}

\date{}

\renewcommand{\headeright}{Weronika Hryniewska-Guzik et al.} 
\renewcommand{\undertitle}{}
\renewcommand{\shorttitle}{\hspace{-2cm}CNN-based explanation ensembling}

\hypersetup{
pdftitle={CNN-based explanation ensembling},
pdfsubject={CNN, XAI},
pdfauthor={Weronika Hryniewska-Guzik et al.},
pdfkeywords={Explainable Artificial Intelligence, Convolutional Neural Network, model evaluation, data evaluation, representation learning, ensemble.},
}

\begin{document}
\maketitle

\begin{abstract}

Explainable Artificial Intelligence has gained significant attention due to~the~widespread use of~complex deep learning models in~high-stake domains such as medicine, finance, and~autonomous cars. However, different explanations often present different aspects of~the model's behavior. In~this research manuscript, we explore the~potential of~ensembling explanations generated by deep classification models using convolutional model.  Through experimentation and~analysis, we aim to~investigate the~implications of~combining explanations to~uncover a~more coherent and~reliable patterns of~the~model's behavior,  leading to~the possibility of~evaluating the~representation learned by the~model. With our method, we can uncover problems of~under-representation of~images in~a~certain class. Moreover, we discuss other side benefits like features' reduction by replacing the~original image with its explanations resulting in~the~removal of~some sensitive information. Through the~use of~carefully selected evaluation metrics from the~Quantus library, we demonstrated the~method's superior performance in~terms of~Localisation and~Faithfulness, compared to~individual explanations.
\end{abstract}


\section{Introduction}
Deep learning models, despite their unprecedented success  \cite{yu2022coca, zhuang2021}, lack full transparency and interpretability in their decision-making processes \cite{Rudin2019, LONGO2024102301}. This has led to~growing concerns about the~use of~"black box" models and~the~need for explanations to~better understand their inferential process~\cite{Poon2021}. Using examples of~specific cases from a~dataset, generated~explanations might reveal which elements are most important in~a model's prediction \cite{sundararajan2017axiomatic, Ribeiro2016, NIPS2017_8a20a862}.

Currently, explanations generated for a~trained deep learning models often are presented as individual insights that need to~be investigated separately and~then compared~\cite{Alber2019}. Each explanation provides a~limited view of~the~model's decision, as it tends to focuse on specific aspects, making it challenging for a~human to~obtain a~comprehensive understanding. This approach hinders the~ability to~discern the~reasons behind a~model's predictions. 

There has been an~emerging trend in~explanation ensembling, which is derived from model ensembling, which involves combining multiple predictive models to~reduce variation of~predictions which often leads to~higher overall performance. Examples of~such predictive techniques are random forests~\cite{Breiman2001} and~gradient boosting~\cite{friedman2001greedy}. This tendency shows that it is plausible that individual explanations possess unique pieces of~information that, when combined, might form a~more comprehensive and~accurate understanding of~a~model's inferential process.~\cite{Zou2022, Bobek2021}

However, despite the~presence of~metrics to~evaluate the~explanations'~\cite{kokhlikyan2020captum, VILONE202189} and~model's~\cite{vujovic2021classification} quality, quantitative metrics that assess the~representation learned by the~model are not well-defined~\cite{10.1609/aaai.v37i10.26392}. Instead, still many research manuscripts~\cite{10035017, 9793621} use various visualisation methods, such as those visualising the components of Principal Component Analysis (PCA)~\cite{kurita2019principal} and~t-Distributed Stochastic Neighbor Embedding (t-SNE)~\cite{van2008visualizing}, which are challenging for quantitative evaluation.

Hence, models continue to~be benchmarked on extensive datasets, like ImageNet~\cite{deng2009imagenet}. Nevertheless, relying solely on single datasets is no longer adequate for effectively distinguishing between models in~terms of~quality~\cite{Beyer2020AreWD}. Evaluating models on benchmarks comprising multiple datasets is recognized as both time and~resource-intensive. However, it is crucial to~note that even such comprehensive evaluations do not ensure an~optimal performance of~a~chosen pretrained model when fine-tuning is applied to~a~specific target dataset \cite{wachinger2016domain}.

For~this reason, our method introduce metrics to~evaluate three important aspects when training representation-based models, namely the~quality of~a~dataset, the~representations learned, and~the~completeness (exhaustiveness) of~explanations. We apply novel ensembling explanations with the~use of~Convolutional Neural Networks (CNNs). One of~ensembling explanations advantages is that they can predict the~possibility of~bias in~both learned representation and~data. This happens when the~results of~training our explanation ensembling method will perform much better on the~training set than on the~testing set. 

In~addition, our approach may be useful for techniques that reduce unnecessary information contained in~the image data, for example by reducing the~number of~features. The removal of sensitive information presented in~images holds particular significance in~critical fields such as medicine to ensure patient privacy~\cite{Aouedi2023, radiol.2020192224, doi:10.1098/rsif.2017.0387}. Furthermore, the~ability to~reduce unnecessary information and~noise from the~original image, potentially might improve training results and speed up it by focusing on relevant features, improving signal-to-noise ratio and reducing risk of overfitting ~\cite{9254002, Jindal2017ARO}. 

Furthemore, our method can be particularly important for classes that might involve sensitive attributes such as race or ethnicity~\cite{Xu2022fairness}. This will help identify the~need to~increase the~number of~images related to a specific class to support their discrimination through model's learning. Another advantage of our method is the~possibility to~check the~impact of~the component explanations on the~final explanation.

Understanding the explanations becomes crucial in this context, especially when they may conflict. This common problem can be overcome with our approach, which can be learned from explanations. It might learn to prioritize one explanation method over another, considering it more reliable. Alternatively, conflicting parts of explanations may lead to a cancellation effect, where our XAI Ensembler decides that certain areas are not relevant.

This manuscript addresses the research question of whether CNN-based explanations aids in dataset, representation, and explanation assessment. In the following sections, we present our approach, introduce metrics and conduct experiments providing insights into the~suitability of our method for evaluation purposes.
\section{Related works}

\paragraph{Explanations for deep learning models} have gained considerable attention in~the~field of~computer vision, particularly in~the~context of~post-hoc analysis. Various approaches have been proposed to~generate meaningful explanations for deep learning models~\cite{kokhlikyan2020captum}. Perturbation-based methods, such as Lime~\cite{Ribeiro2016}, Occlusion~\cite{zeiler2013visualizing}, and~Shapley Value Sampling~\cite{CASTRO20091726}, involve modifying an~input image to~evaluate the~impact on a~model's output and~identify significant regions or features. On the~other hand, gradient-based methods, including Integrated Gradients~\cite{sundararajan2017axiomatic}, DeepLift~\cite{10.5555/3305890.3306006}, Guided GradCAM~\cite{Selvaraju_2019}, GradientShap~\cite{NIPS2017_8a20a862}, or Guided Backpropagation~\cite{SpringenbergDBR14}, utilize gradients to~highlight the~salient regions of~an~input image that contribute most to~the~model's inferential process. Additionally, Noise Tunnel~\cite{10.5555/3454287.3455160} is an~approach that introduces noise into the~input image and~analyzes its impact on the~model's output. 

\paragraph{The~ensemble of~explanations} has not yet received significant attention in~the~field of~Explainable Artificial Intelligence (XAI). Zou et al.~\cite{Zou2022} proposed an~approach for XAI ensembling using a~Kernel Ridge regression~\cite{10.5555/2380985} to~combine the~normalized Grad-CAM++ technique~\cite{Selvaraju_2019} and~the~normalized positive SHAP values. In~an~alternative method for XAI ensembling based on metrics (consistency, stability, and~area under the~loss)~\cite{Bobek2021}, the~authors calculated the~weighted importance of~three different metrics  and~created a~combined vector of~explanations by taking a~weighted sum of~ensemble scores and~their corresponding original explanations. Another study explored the~use of~mean to~combine explanations generated by multiple explanation methods~\cite{Rieger2020}, aiming to~provide a~more noise-resistant ensembled explanation. Additionally,  Bhatt et al.~\cite{Bhatt2020} proposed a~XAI ensembling method based on optimization around the~points of~interest, minimizing explanation sensitivity and~bridging the~gap between AI decision-making system and~human interpretability, particularly in~the~context of~medical applications. These research works contribute to~the~growing body of~research on XAI ensembling, offering various techniques and~perspectives to~enhance the~interpretability and~trustworthiness of~AI systems. It is worth mentioning that ensembles of~explanations also appear in~the~literature under the~title of~aggregations of~explanations \cite{Bhatt2020, Rieger2020}.

Furthermore, NoiseGrad \cite{bykov2022noisegrad} and SmoothGrad \cite{smilkov2017smoothgrad} explanation methods can be considered as ensembles of explanations. NoiseGrad injects noise into the model's weights and then averages over perturbed models. At the same time, SmoothGrad adds noise to the input data and then averages the resulting heatmaps for each of the~input images.

\paragraph{Evaluation of~explanations} using pixel-wise masks plays a~crucial role to~assess the~quality and~reliability of~computer vision models. Several works have proposed metrics and~approaches to~evaluate explanations using pixel-wise masks. One of~such contributions is the~benchmark dataset CLEVR-XAI~\cite{Arras2022}, which was specifically designed for the evaluation of neural network explanations with ground truth. The~authors proposed two novel quantitative metrics, relevance mass accuracy and~relevance rank accuracy, which are suitable for evaluating visual explanations in~reference to~ground-truth masks. They emphasized the~importance of~establishing a~well-defined common ground for evaluating XAI methods, which can significantly advance XAI research. To address challenges associated with the~assumption that humans can accurately recognize explanations~\cite{wang2019,Rosenfeld2021}, the~Quantus library~\cite{Hedstrom2022} was developed. This library offers a~versatile and~comprehensive toolkit for collecting, organizing, and~explaining various evaluation metrics proposed for explanation methods. It aims to~automate the~process of~explanation quality quantification, assisting researchers in~the~evaluation of~explanation techniques, including localization metrics. 

In~the~work~\cite{Zhang2018}, Excitation Backprop was introduced as a~technique for evaluating explanations. The~evaluation process involved verifying whether the~single pixel with the~highest relevance lies within the~object corresponding to~a~model's predicted class, referred to~as the~pointing game accuracy. This metric assesses the~ability of~an~explanation method to~accurately highlight the~salient regions associated with the~predicted class by the underlying trained model. These notable research contributions highlight the~importance of~developing rigorous evaluation methodologies for explanation techniques in~the~context of~pixel-wise masks. The~proposed metrics and~tools provide valuable resources to~the~XAI community, enabling researchers to~objectively assess and~compare the~performance of~different explanation methods in~terms of~their localization accuracy and~relevance to~ground truth masks.

\paragraph{Representation learning} aims to~distill complex data into low-dimensional, reusable structures that exhibit specific qualities~\cite{oh2020deepmicro}. A~good representation is characterized by several key attributes. These include low dimensionality for efficiency, spatial coherence for proximity of~related aspects, and~disentanglement to~isolate specific features~\cite{Lin_2019_CVPR_Workshops}. Additionally, maintaining hierarchy and~meaningfulness aids interpretability.

The~utilization of~models that have informative representations facilitates the~generation of~segmentation masks through explanations. This approach has gained popularity in~weakly-supervised semantic segmentation problems~\cite{Liu2022, Chen_2022_CVPR, Chen_2022_CVPR_b}. Typically, these studies employ the~Class Activation Mapping (CAM) explanation method to~provide explanations and~then to~generate segmentation masks. However, recent work~\cite{Saporta2022} demonstrates that other explanation methods can also be used to~obtain pixel-wise masks.

In summary, while post-hoc explanations for deep learning models have seen significant advancements, the potential of generating masks using explanations ensemble methods remains unexplored. To the best of our knowledge, there are no metrics for the evaluation of explanations that are based on the relevance of the component explanations in the ensemble of explanations. In the upcoming section, we delve into details of the method and conduct experiments to explore the effectiveness of our CNN-based explanations ensembling.

\paragraph{Evaluation of representation} goes beyond individual image analysis using post-hoc explainability methods. Moreover, relying solely on the~accuracy on ImageNet may be misleading, as a high accuracy metric does not guarantee effective representation learning or generalization. Beyer et al.~\cite{Beyer2020AreWD} emphasize that performance metrics on large datasets may not capture a~model's ability to~handle diverse data. Beyond accuracy, evaluating representations through interpretability methods or transfer learning tasks provides a~more comprehensive measure of~a~model's capabilities, ensuring it generalizes effectively in~real-world scenarios.

While t-SNE~\cite{van2008visualizing} is commonly chosen to~compare learned representations visually~\cite{s21238070, 8682397}, it has limitations. Firstly, t-SNE effectively preserves local structure but may not accurately represent global relationships in~the data. It tends to~focus on preserving pairwise similarities between nearby data points, which may not capture the~overall structure of~the data. Secondly, t-SNE has hyperparameters, such as the~perplexity, that need to be set, which can significantly impact the~resulting visualization. Selecting appropriate values for these hyperparameters can be subjective and~may lead to~different interpretations of~the data. Lastly, t-SNE can be sensitive to~noisy or sparse data, potentially producing misleading visualizations. In~the~context of~evaluating deep learning representations, noisy or irrelevant features may influence the~t-SNE visualization in~ways that do not accurately reflect the~quality of~the learned representations. 

To conclude, t-SNE can be a~powerful tool for visualization, but it fails in~quantitative assessment of~data representation quality. Additionally, precision, recall, f1 score and auc roc are good metrics for assessing the~quality of~classification on a~given dataset. However, they are not suitable for assessing the~quality of~representation, because they do not check the image features learned by a~model.


\section{The~concept of~CNN-based ensembled explanations}

\begin{figure*}[h]
    \centering
    \includegraphics[width=\textwidth]{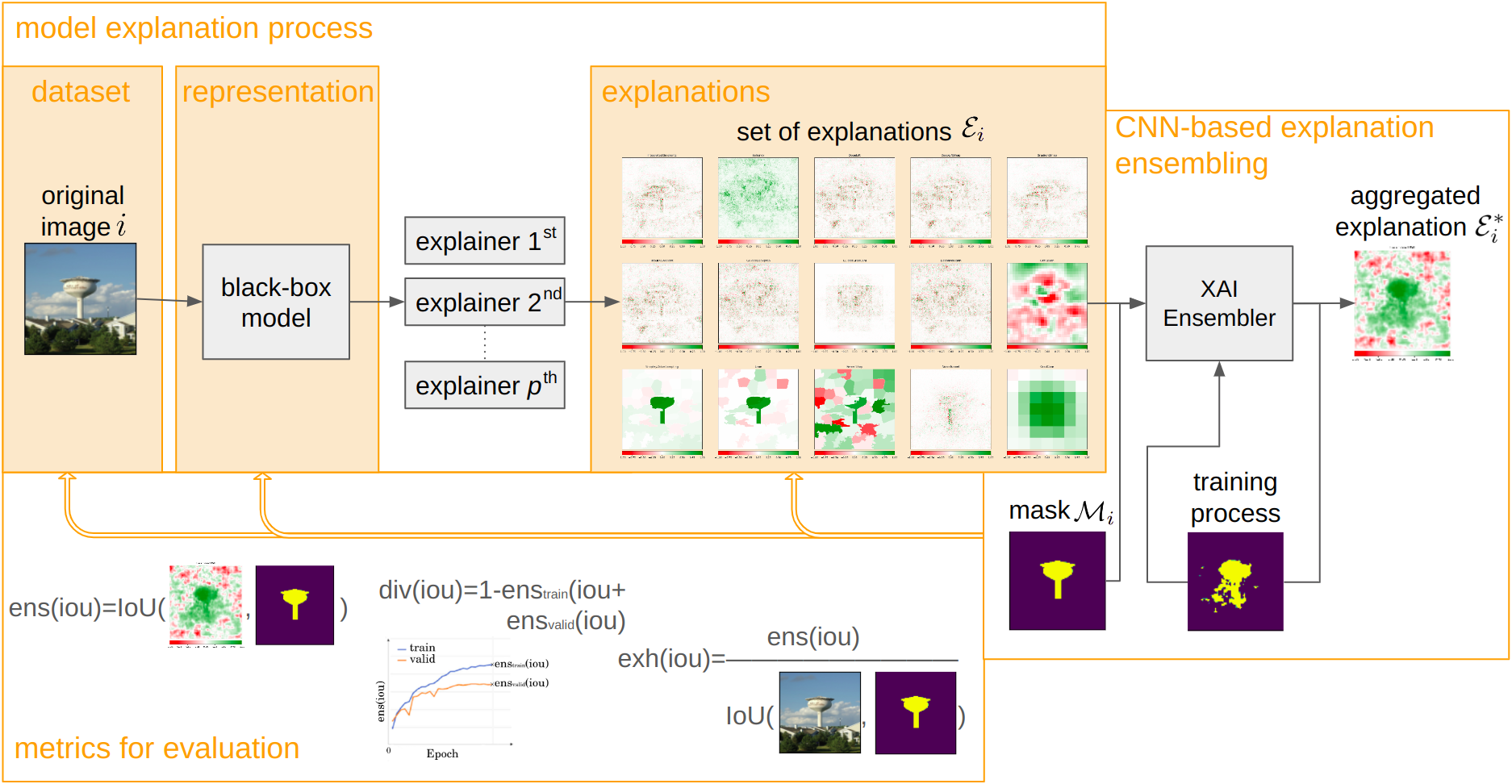}
    \caption{Overview of~CNN-based ensembling of~explanations. The~diagram illustrates the~process that involves an~original image being processed by a~black-box classification model, followed by an~explainer generating explanations. These explanations, along with masks depicting objects in~the~original image, are then used to~train XAI Ensembler, which architecture is presented in Figure \ref{fig:architecture}. The~ensembler's training process results in~an~aggregated image. CNN-based ensembling of~explanations with metrics we defined can be used to assess dataset, representation and explanations.}
    \label{fig:mainidea}
\end{figure*}

In~our proposed methodology, presented in~Figure \ref{fig:mainidea} and~Algorithm \ref{alg:segmentation_ensemble}, we introduce a~novel approach for ensembling explanations to~enhance the~interpretability of~classification models. The~process begins by generating explanations using established techniques for the~trained classification model. Consider a~set of~$p$ explanation methods. Let \(\mathcal {E}_{ij}\) represent the~explanation created by~\(j\)-th method for the~\(i\)-th image (\(I_i\)) in~the~dataset. For each image, we aim to~generate a~set of~explanations:

\begin{equation}
\mathcal {E}_{i} = \{\mathcal {E}_{i1}, \mathcal {E}_{i2}, \ldots, \mathcal {E}_{ip}\}.
\end{equation}

To ensemble these explanations, we employ two strategies depending on the~availability of~ground truth masks. Let \(\mathcal {M}_i\) denote the~pixel-wise annotated mask corresponding to~the~\(i\)-th image. In~the~first strategy, we utilize pixel-wise annotated masks for a~selected class of~images, which serve as a~reference for generating ensembles. Alternatively, in~cases where annotated masks are not readily available, a domain specialist can manually annotate a~small subset of~representative images.

Once the~explanations are generated, our objective is to~train a~CNN-based explanation ensembling model (XAI Ensembler), shown in~Figure \ref{fig:architecture}, to~predict masks given explanations. The~training procedure involves minimizing the~segmentation loss \(\mathcal {L}_{\text{seg}}\) as follows:

\begin{equation}
\mathcal {L}_{\text{seg}} = \sum_{i=1}^{n} \text{SoftDice}(\mathcal {M}_i, \text{XAI\_Ensembler}(\mathcal {E}_i)),
\end{equation}

where \(n\) represents the~total number of~images. The~Soft Dice coefficient~\cite{soft}, refers to~a~similarity metric that quantifies the~overlap between predicted and~ground-truth segmentation masks and~is utilized to~assess the~segmentation accuracy of~the \textit{XAI Ensembler} model. The~explanation aggregation obtained by XAI Ensembler is denoted as $\mathcal {E}^*_i$.

The~detailed training procedure for this segmentation model is outlined in~detail in~Section \ref{trainingprocedure}. It provides comprehensive insights into the~specific steps, techniques, and~parameters employed during the~training process. We aim to~train a~segmentation model that effectively captures and~interprets the~underlying image features by incorporating the~generated explanations and~their corresponding masks.

After training the~segmentation model using the~explanations and~masks, we progress to~evaluating its performance with unseen images. This evaluation serves as a~crucial step in~assessing the~generalization capabilities and~effectiveness of~the~trained segmentation model. By applying the~model to~unseen data, we can gain valuable insights into its ability to~accurately delineate the~regions of~interest and~align with human perception.

Overall, our proposed method of~ensembling explanations offers a~unique perspective on interpreting and~understanding the~predictions of~a~trained classification model. Through the~integration of~explanations and~the~subsequent training of~a~segmentation model, we strive to~enhance the~interpretability and~trustworthiness of~the~underlying classification model. The~evaluation of~the~trained segmentation model with unseen images serves as a~crucial testing step, providing valuable insights into the~complex classification model.

\begin{algorithm}
\caption{CNN-based Ensembled Explanations training}
\label{alg:segmentation_ensemble}
\begin{algorithmic}[1]
\IF{pixel-wise masks available}
    \STATE Utilize pixel-wise annotated masks $\mathcal {M}$ for ensembling.
\ELSE
    \STATE Manually annotate a~small subset of~representative images.
\ENDIF
\STATE Select the~classification model for which the~ensemble of~explanations will be generated.
\STATE Choose the~number of~epochs $e$.
\STATE Count total number of~images $n$.
\STATE Select $p$ explanation methods for ensembling.
\STATE Define the~architecture of~the~CNN-based explanation ensembling model
\FOR{$d$ = 1 to~$e$}
\FOR{$i$ = 1 to~$n$}
\FOR{$j$ = 1 to~$p$}
\STATE Generate explanation $\mathcal {E}_{ij}$ for the~trained classification model.
\ENDFOR

\STATE Train the~CNN-based explanation ensembling model using the~explanations $\mathcal {E}_i$ and masks $\mathcal {M}_i$ (see Section~\ref{trainingprocedure}).
\ENDFOR
\ENDFOR
\end{algorithmic}
\end{algorithm}

\subsection{Experimental setup for training}\label{trainingprocedure}

Our experiments employed a~segmentation model based on CE-Net~\cite{8662594}, which has shown promising performance in~various image analysis tasks \cite{hryniewskaguzik2024comparative}. The~dataset used for training and~evaluation was ImageNet-S50~\cite{Deng2010}. We split the data 80/20 for training and testing. The~input images were resized to~224x224 pixels. During training, we utilized a~learning rate of~0.0002, and~the~learning rate was halved when no improvement was observed, with a~minimum interval of~20 epochs. The~training process was stopped if the~learning rate dropped below $10^{-9}$, or after 200 epochs.

To optimize the~segmentation model, we employed the~Soft dice loss~\cite{soft} and~utilized the~Adam optimizer~\cite{adam}. The~Soft dice loss has been proven effective in~handling classes with lesser spatial representation in an image, and~is commonly used in~medical image segmentation tasks~\cite{9116807}.

To enhance the~robustness of~the~model and~minimize overfitting, we applied various image transformations, which are standard for ImageNet. These included a~center crop to~focus on the~central region of~the~image, Imagenet normalization to~standardize pixel values, random adjustments of~hue and~saturation values, random shifts, scaling, and~rotations, random flips to~introduce horizontal and~vertical symmetry variations, as well as random rotations by 90°.

The~code written using PyTorch library is available on Github: \url{https://github.com/Hryniewska/CNNbasedXAIensembling}.


\subsection{Ablation studies}

\begin{figure*}[h]
  \centering

  \begin{subfigure}{\textwidth}
  \resizebox{\textwidth}{!}{
    
    \begin{tikzpicture}[
      conv/.style={draw, rounded corners, fill=blue!20, minimum width=1cm, minimum height=1cm},
      bn/.style={draw, rounded corners,fill=orange!20, minimum width=0.7cm, minimum height=1cm},
      relu/.style={draw, rounded corners,fill=yellow!20, minimum width=1.0cm, minimum height=1cm},
      pool/.style={draw, rounded corners,fill=orange!20, minimum width=0.7cm, minimum height=1cm},
      deconv/.style={draw, rounded corners,fill=green!20, minimum width=1.0cm, minimum height=1cm},
      resnet/.style={draw, rounded corners,fill=blue!20, minimum width=1.5cm, minimum height=1cm},
      concat/.style={draw, rounded corners,fill=magenta!20, minimum width=1cm, minimum height=1cm},
      sum/.style={draw, rounded corners,fill=magenta!20, minimum width=1cm, minimum height=1cm},
      sigmoid/.style={draw, rounded corners,fill=yellow!20, minimum width=1.0cm, minimum height=1cm},
      arrow/.style={thick,->,>=stealth}
      ]

      \node[conv] (conv1_1) {Conv};
      \node[bn, right=0.5cm of conv1_1] (bn1_1) {BN};
      \node[relu, right=0.5cm of bn1_1] (relu1_1) {ReLU};
      \node[pool, right=0.5cm of relu1_1] (pool1_1) {MaxP};
      \node[resnet, right=0.5cm of pool1_1] (resnet1_1) {ResNet(1)};
      \node[resnet, right=0.5cm of resnet1_1] (resnet2_1) {ResNet(2)};
      \node[resnet, right=0.5cm of resnet2_1] (resnet3_1) {ResNet(3)};
      \node[resnet, right=0.5cm of resnet3_1] (resnet4_1) {ResNet(4)};

      \node[conv, below=1cm of conv1_1] (conv1_2) {Conv};
      \node[bn, right=0.5cm of conv1_2] (bn1_2) {BN};
      \node[relu, right=0.5cm of bn1_2] (relu1_2) {ReLU};
      \node[pool, right=0.5cm of relu1_2] (pool1_2) {MaxP};
      \node[resnet, right=0.5cm of pool1_2] (resnet1_2) {ResNet(1)};
      \node[resnet, right=0.5cm of resnet1_2] (resnet2_2) {ResNet(2)};
      \node[resnet, right=0.5cm of resnet2_2] (resnet3_2) {ResNet(3)};
      \node[resnet, right=0.5cm of resnet3_2] (resnet4_2) {ResNet(4)};

      \node[conv, below=1cm of conv1_2] (conv1_3) {Conv};
      \node[bn, right=0.5cm of conv1_3] (bn1_3) {BN};
      \node[relu, right=0.5cm of bn1_3] (relu1_3) {ReLU};
      \node[pool, right=0.5cm of relu1_3] (pool1_3) {MaxP};
      \node[resnet, right=0.5cm of pool1_3] (resnet1_3) {ResNet(1)};
      \node[resnet, right=0.5cm of resnet1_3] (resnet2_3) {ResNet(2)};
      \node[resnet, right=0.5cm of resnet2_3] (resnet3_3) {ResNet(3)};
      \node[resnet, right=0.5cm of resnet3_3] (resnet4_3) {ResNet(4)};      

      \node[conv, right=1cm of resnet4_2] (dac) {DAC};
      \node[conv, right=0.5cm of dac] (rmp) {RMP};

      \node[concat, above=1cm of dac] (concat4) {Concat / Sum};
      \node[concat, above=0.5cm of concat4] (concat3) {Concat / Sum / SkipCon};
      \node[concat, above=0.5cm of concat3] (concat2) {Concat / Sum / SkipCon};
      \node[concat, above=0.5cm of concat2] (concat1) {Concat / Sum / SkipCon};

      \node[deconv, right=0.5cm of rmp] (deconv4) {Decoder4};
      \node[deconv, right=0.5cm of deconv4] (deconv3) {Decoder3};
      \node[deconv, right=0.5cm of deconv3] (deconv2) {Decoder2};
      \node[deconv, right=0.5cm of deconv2] (deconv1) {Decoder1};

      \node[sum, above=0.5cm of deconv4] (sum4) {Sum};
      \node[sum, above=0.5cm of deconv3] (sum3) {Sum};
      \node[sum, above=0.5cm of deconv2] (sum2) {Sum};

      \node[deconv, right=0.5cm of deconv1] (deconv_final) {Deconv};
      \node[relu, right=0.5cm of deconv_final] (relu_final) {ReLU};
      \node[conv, right=0.5cm of relu_final] (conv_final) {Conv};
      \node[sigmoid, right=0.5cm of conv_final] (sigmoid) {Sigm};

     \draw [decorate, decoration = {brace,raise=30pt,amplitude=5pt}] ([xshift=-1cm]conv1_3.south) -- ([xshift=-1cm]conv1_1.north)  node[black,pos=0.5,left=40pt, text width=3cm, font=\Large] {number of encoders};

    \node (foo) [above=50pt, left=10pt, font=\Large] at (2.5,1.5) {input with number of channels};
    
    \draw [arrow] (foo) |- (conv1_1);  
    \draw [arrow] (foo) |- (conv1_2);  
    \draw [arrow] (foo) |- (conv1_3);  

      
      \draw[arrow] (conv1_1) -- (bn1_1);
      \draw[arrow] (bn1_1) -- (relu1_1);
      \draw[arrow] (relu1_1) -- (pool1_1);
      \draw[arrow] (pool1_1) -- (resnet1_1);
      \draw[arrow] (resnet1_1) -- (resnet2_1);
      \draw[arrow] (resnet2_1) -- (resnet3_1);
      \draw[arrow] (resnet3_1) -- (resnet4_1);

      \draw[arrow] (conv1_2) -- (bn1_2);
      \draw[arrow] (bn1_2) -- (relu1_2);
      \draw[arrow] (relu1_2) -- (pool1_2);
      \draw[arrow] (pool1_2) -- (resnet1_2);
      \draw[arrow] (resnet1_2) -- (resnet2_2);
      \draw[arrow] (resnet2_2) -- (resnet3_2);
      \draw[arrow] (resnet3_2) -- (resnet4_2);

      \draw[arrow] (conv1_3) -- (bn1_3);
      \draw[arrow] (bn1_3) -- (relu1_3);
      \draw[arrow] (relu1_3) -- (pool1_3);
      \draw[arrow] (pool1_3) -- (resnet1_3);
      \draw[arrow] (resnet1_3) -- (resnet2_3);
      \draw[arrow] (resnet2_3) -- (resnet3_3);
      \draw[arrow] (resnet3_3) -- (resnet4_3);

      \draw[arrow] (resnet1_1) -- ++(1.0,1cm) |- (concat1);
      \draw[arrow] (resnet1_2) -- ++(1.0,1cm) |- (concat1);
      \draw[arrow] (resnet1_3) -- ++(1.0,1cm) |- (concat1);

      \draw[arrow] (resnet2_1) -- ++(1.0,1cm) |- (concat2);
      \draw[arrow] (resnet2_2) -- ++(1.0,1cm) |- (concat2);
      \draw[arrow] (resnet2_3) -- ++(1.0,1cm) |- (concat2);

      \draw[arrow] (resnet3_1) -- ++(1.0,1cm) |- (concat3);
      \draw[arrow] (resnet3_2) -- ++(1.0,1cm) |- (concat3);
      \draw[arrow] (resnet3_3) -- ++(1.0,1cm) |- (concat3);

      \draw[arrow] (resnet4_1) -- ++(1.0,1cm) |- (concat4);
      \draw[arrow] (resnet4_2) -- ++(1.0,1cm) |- (concat4);
      \draw[arrow] (resnet4_3) -- ++(1.0,1cm) |- (concat4);

      \draw[arrow] (concat4) -- (dac);
      
      \draw[arrow] (dac) -- (rmp);

       \draw[arrow] (rmp) -- (deconv4);
       \draw[arrow] (deconv4) -- (sum4);
       \draw[arrow] (deconv3) -- (sum3);
       \draw[arrow] (deconv2) -- (sum2);

       \draw[arrow] (sum4) -- (deconv3);
       \draw[arrow] (sum3) -- (deconv2);
       \draw[arrow] (sum2) -- (deconv1);

       \draw[arrow] (concat3)  -- ++(2,0cm) -| (sum4);
       \draw[arrow] (concat2) -- ++(2,0cm) -|  (sum3);
       \draw[arrow] (concat1)  -- ++(2,0cm) -| (sum2);

       \draw[arrow] (deconv4) -- (sum4);
       \draw[arrow] (deconv3) -- (sum3);
       \draw[arrow] (deconv2) -- (sum2);

      \draw[arrow] (deconv1) -- (deconv_final);
      \draw[arrow] (deconv_final) -- (relu_final);
      \draw[arrow] (relu_final) -- (conv_final);
      \draw[arrow] (conv_final) -- (sigmoid);

    \end{tikzpicture}%
    \newline
  }
  \includegraphics[width=0.965\textwidth, right]{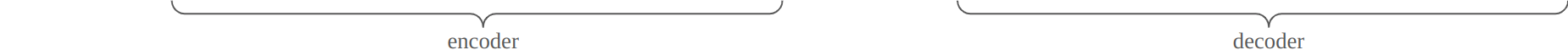}
  \end{subfigure}
  
  \caption{Proposed three architectures of CNN-based explanation ensembling: concatenation, summation and multi-channel approach. Key architectural components are abbreviated as follows: BN (BatchNorm2D), MaxP (MaxPool2D), Conv (Conv2D), Concat (Concatenation), Sum (Summation), and SkipCon (Skip connection). The~number in~brackets after "ResNet" denotes the~specific number of layers in~the~ResNet architecture.}
    \label{fig:architecture}
\end{figure*}
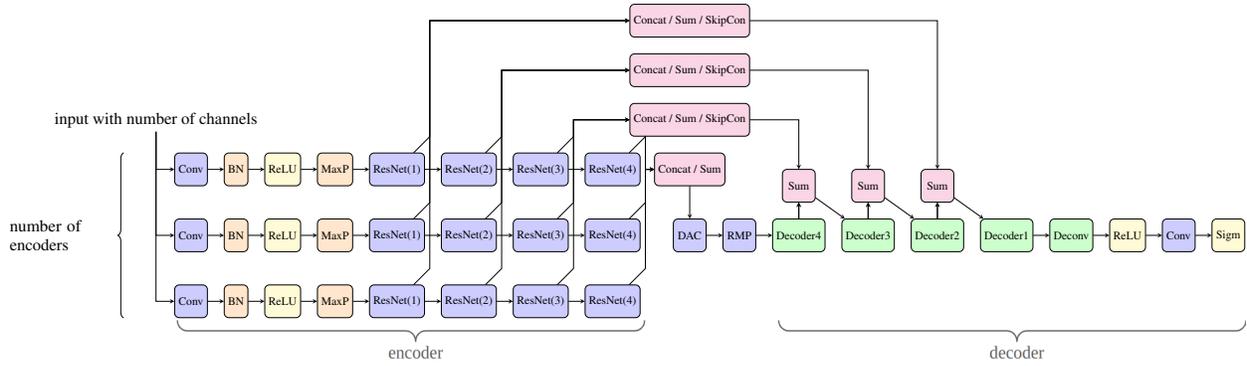

The~objective of~our experiments is to~select the~best ensemble model's architecture. We created three different model's architectures, presented in~Figure~\ref{fig:architecture}. The~first and~the~second architecture used many encoders. Each encoder takes one explanation method as an~input; therefore the~number of~encoders is linearly dependent on the~number of~explanations that are going to~be ensembled.

In~the~first architecture, we used concatenation as the~method for combining inputs from multiple encoders. Concatenation involves stacking encoders' outputs along a~new channel dimension. This type of stacking is useful when you want to~retain all the~information from each input and~allow the~network to~learn how to~use each input separately. It is particularly suitable for problems where the~inputs represent different aspects of~the problem. It is worth noting that the~dimensionality of~the input space increases when using concatenation, potentially leading to~a~large number of model parameters. This architecture's total number of parameters is 17 675 256 multiplied by the number of encoders squared.

The~second approach employed summation as the~combining method. It entails an~element-wise addition of~the encoder outputs, channel by channel. Summation is used when the~inputs are expected to~be complementary, and~it emphasizes their joint influence on the~output. This method helps keep the~model's number of~parameters low (17 675 256 parameters) and~can be beneficial when the~inputs are additive.

The~last architecture consists of~a~single encoder that takes explanations in~channels, resulting in~the number of~channels being a~product of~three and~the~number of~explanations for ensembling. The Captum library~\cite{kokhlikyan2020captum}, employed for generating explanations, provides output with three channels for each explanation. The output values fall within the range of [-1, 1] or [0, 1], depending on the explanation method. Additional channels in~convolutional layers are appropriate when we need to~leverage the~spatial relationships in~the inputs while allowing the~network to~learn features from each input independently. It is commonly used in~tasks like multi-modal image processing, where different inputs (RGB image, depth map, infrared image) are processed jointly to~improve feature extraction~\cite{li2019learning}. The number of parameters is equal to 17 675 256.

Typically, the~choice among these methods depends on empirical experimentation and~the~specific requirements of~a~machine learning or deep learning task. It is important to~consider the~unique characteristics of~the data and~the~problem to~determine which input combination method is most appropriate.

For this reason, we trained all three architectures using the~same training procedure, as described in~Section \ref{trainingprocedure}. The~proposed model architectures were trained on several dozen models with the~same configurations encompassing various explained deep learning classification models, explanation methods, and~specific classes. These configurations included explained deep learning classification models, such as SqueezeNet1.1 MobileNet v2, VGG16, DenseNet121, ResNet50, and~EfficientNet B0. An effort was made to select diverse classes, e.g. objects that are large or small, popular or rare, always look the same or always differently. The target classes comprised digital watch, dining table, gibbon (monkey), grand piano, kuvasz (dog), ladybird, purse, umbrella, beach wagon, water bottle, and water tower. The~experimental setup for sets of~explanation methods consisted of~the~following configurations: zero explanations (the original image is displayed instead of~the~explanation), three local explanations that are based on different approaches (Shapley Value Sampling,  NoiseTunnel,  Guided Backprop), four explanations based on the~most frequently cited XAI methods (GradientShap,  Lime,  Occlusion, Guided GradCAM) and~seven~explanations that show differentiation between XAI methods (DeepLift or Integrated Gradients for ResNet,  GradientShap,  Guided Backprop,  Guided GradCAM,  Lime,  Occlusion, Shapley Value Sampling).

The~obtained results were averaged and~are presented in~Table~\ref{tab:architecture}. The detailed interpretation of the used metrics is presented in Section \ref{sec:metrics}. However, it can be seen that the best results are obtained for the architecture with the concatenation operation, so all further experiments will only be performed using this architecture.

\begin{table*}[h]
    \centering
    \setlength{\tabcolsep}{2pt}
    \begin{tabular}{@{}p{2cm}|rrrr|rrrr@{}}
        \textbf{Ensembling} & \multicolumn{4}{c|}{\textbf{Training}} & \multicolumn{4}{c}{\textbf{Testing}} \\
        \textbf{method} & \multicolumn{1}{c}{\textbf{ens(acc)}} & \multicolumn{1}{c}{\textbf{ens(f1)}} & \multicolumn{1}{c}{\textbf{ens(IoU)}} & \multicolumn{1}{c|}{\textbf{loss}} & \multicolumn{1}{c}{\textbf{ens(acc)}} & \multicolumn{1}{c}{\textbf{ens(f1)}} & \multicolumn{1}{c}{\textbf{ens(IoU)}} & \multicolumn{1}{c}{\textbf{loss}} \\ 
        \midrule
        Concatenation & 0,930 & 0,828 & 0,734 & 0,314 & 0,865 & 0,723 & 0,599 & 0,405 \\
        Sum & 0,898 & 0,615 & 0,504 & 0,233 & 0,809 & 0,511 & 0,407 & 0,379 \\
        Channel & 0,881 & 0,546 & 0,440 & 0,271 & 0,792 & 0,425 & 0,332 & 0,407 \\
        \bottomrule
    \end{tabular}
    \caption{Performance metrics of~different ensembling model architectures on training and~testing sets.}
    \label{tab:architecture}
\end{table*}

\subsection{Method evaluation}

The~decision to~employ the~Quantus library for the~evaluation of~our CNN-based XAI explanation ensembling model is rooted in~its capability to~offer a~comprehensive and~systematic comparison of~various XAI methods. Quantus~\cite{Hedstrom2022} provides a~library for evaluating the~performance of~XAI techniques, ensuring a~thorough examination of~key aspects.

The~chosen metrics for this evaluation were thoughtfully selected to~optimize the~meta-consistency score~\cite{hedstrom2023metaquantus}. It is pivotal as it indicates the~stability and~reliability of~an~XAI method across different contexts and~datasets. Prioritizing metrics contributing to~a~higher meta-consistency score ensures that our method not only explains individual instances effectively but also maintains consistency across diverse scenarios.

\begin{figure}[h]
    \centering
     \includegraphics[width=0.8\linewidth]{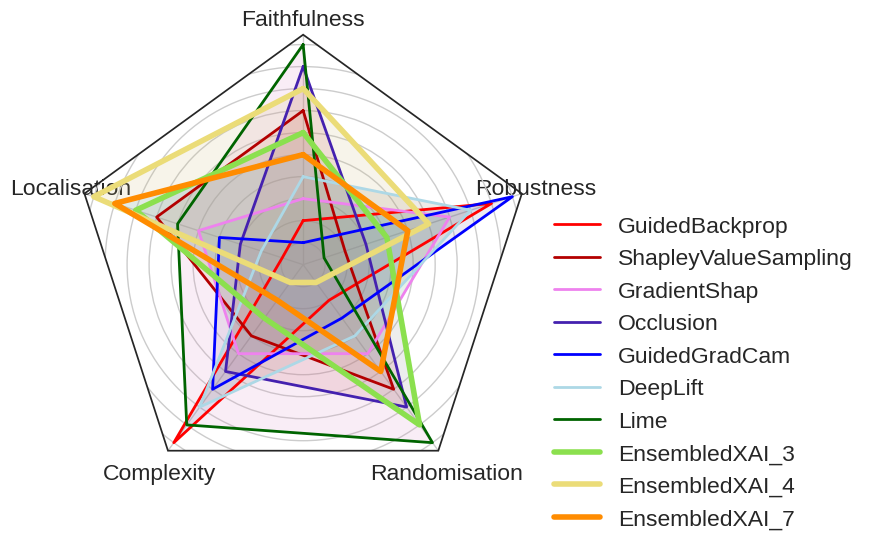}
    \caption{Radar plot comparing the~performance of~CNN-based explanation ensembling with individual non-ensembled XAI methods. The~plot displays the~ranking of explanations in~five key evaluation metrics: Faithfulness, Robustness, Localisation, Complexity, and~Randomisation. Higher values indicate better performance.}
    \label{fig:quantus}
\end{figure}

The~radar plot in~Figure \ref{fig:quantus} visually represents the~comparative performance of~our CNN-based XAI ensembling method against individual non-ensembled XAI methods. The~inclusion of~key evaluation metrics—Faithfulness, Robustness, Localisation, Complexity, and~Randomisation—offers a~holistic view of~a~method's capabilities. The subsequent paragraphs delve into detailed explanations of each of these five metrics.

\paragraph{Localisation (Pointing-Game~\cite{Zhang2018})} This metric helps to evaluate whether the~highest-scored attribution in~the~explanation is accurately localized within the~targeted object or region of~interest. It provides valuable insights into the~accuracy and~precision of~the~explanations in~pinpointing relevant features.
In~terms of~Localisation, CNN-based explanation ensembling method exhibited the~best performance over all explanations.

\paragraph{Faithfulness (Pixel-Flipping~\cite{bach})} This metric assesses the~faithfulness of~the~explanations by perturbing pixels in~descending order of~their attributed values and~observing the~impact on the~classification score. The results are aggregated using mean. Our CNN-based explanation ensembling method achieved competitive scores in~the~Faithfulness metric, ranking near the~top.

\paragraph{Robustness (Local Lipschitz Estimate~\cite{alvarezmelis2018robustness})} This metric measures the~consistency of~explanations between adjacent examples. It quantifies how much the~explanations change when the~input data points are slightly perturbed, providing insights into the~stability and~reliability of~the~explanations. Our CNN-based explanation ensembling method demonstrated relatively good robustness, ranking in the middle of the other selected XAI methods.

\paragraph{Complexity (Sparseness~\cite{chalasani20a})} This metric utilizes the~Gini index to~measure the~sparseness of~the~explanations. It helps assess whether an~explanation predominantly highlights highly attributed features that are truly predictive of~a~model's output. The~complexity metric favors explanations that fill a~small part of~the~segmentation mask. The~principle of~our method does not allow for high scores on this metric.  In our approach, the explanation should contain as much of the mask as possible. However, during the XAI Ensembler model's inference, the cut-off can be increased to get smaller areas marked.

\paragraph{Randomisation (Model Parameter Randomisation~\cite{Adebayo})} While MetaQuantus~\cite{hedstrom2023metaquantus} recommends Random Logit for randomizing model parameters, in~evaluation of our XAI ensembling models, we used an~alternative approach due to the impossibility of carrying out calculations for a random other class. The~ModelParameterRandomisation metric involves randomizing the~parameters of~single model layers in~a~cascading or independent manner. It then measures the~distance between the~original and~randomized explanations, providing insights into the~robustness of~the~original explanations. The results are aggregated using mean.  In~the~Randomisation metric, our CNN-based explanation ensembling method achieved relatively competitive scores, particularly when ensembled with the~three XAI methods, outperforming GuidedBackprop and~GuidedGradCam. However, suprisingly, the~scores associated to our ensembling with 4 XAI methods were low, which may be related to the component explanations chosen.

By utilizing the~Quantus library and~the~metrics following MetaQuantus~\cite{hedstrom2023metaquantus} recommendations, we were able to~systematically evaluate and~compare the~performance of~our CNN-based explanation ensembling. Higher values on the radar plot signify superior performance in these metrics. Our XAI method may excel in faithfully representing the underlying model, demonstrating potential robustness across diverse scenarios. Moreover, it may accurately localise influential features, manage complexity effectively, and exhibit controlled randomization. However, our approach results in~lower scores in~the~complexity metric. 

This evaluation approach emphasizes the~qualitative aspects of~the proposed solution, going beyond a~mere quantitative analysis. By aligning with best practices in~XAI evaluation, the~use of~Quantus metrics may help enhancing the~credibility and~reliability of~our XAI ensemble method.

\section{Metrics for representation, dataset and~explanation evaluation}\label{sec:metrics}

While post-hoc explainability methods provide insights into individual images, they lack a holistic view. These methods fail to comprehensively assess factors like dataset representativeness, learned representations, and explanation quality. Furthermore, relying solely on ImageNet accuracy can be misleading. Even visualization techniques like t-SNE fail in~quantitative assessment of~data representation quality.

\textbf{Representation-oriented ensembling performance metric}
\nopagebreak

To address these challenges, a~novel metric is proposed for quantitatively assessing data representation quality through CNN-based XAI ensembling. The~objective of~explanations is to~reveal the~elements within an~image that guide the~model's inferential process. Evaluating these explanations becomes challenging when multiple methods generate diverse interpretations \cite{krishna2022disagreement}. This is where CNN-based XAI ensembling could prove valuable. The~model endeavors to~learn to~segment the~object by exclusively considering explanations and~the~ground-truth mask of~the classified object. Success in~learning indicates that explanations might effectively represent the~classified object, minimizing the~impact of~noise for an~accurate interpretation.

The~matrix $\mathcal{E}_i$, defined in values within the~range $[-1,1]$. To compute the~evaluation metrics for \textit{representation-oriented ensembling performance (ens)}, we discretize $\mathcal{E}_i$ into a~binary version denoted as $\mathcal{E}^*_{bin}$, where values above a~selected cut-off are set to~1, and~values below are set to~0. In~this research work, we use cut-off equal to~0.5, which is typical for segmentation tasks. The~evaluation metric ens(iou), which measures the~overlap between the~one values of~$\mathcal{E}_{bin}^*$ and~any overlapping ground-truth instances $\mathcal{M}$, is calculated as follows:

\begin{equation}
ens(iou) = IoU(\mathcal{E}_{bin}^*, \mathcal{M}).
\end{equation}

Other metrics for CNN-based XAI ensembling, such as ens(f1) or ens(acc), can be calculated analogously by comparing the binary ensemble of explanations~($\mathcal{E}_{bin}^*$) with the ground truth ($\mathcal{M}$). True positives, true negatives, false positives and false negatives are determined based on the overlap between the binarised explanations and the ground truth.

\textbf{Data-oriented diverseness metric}
\nopagebreak

In scenarios, when an image contains multiple objects (for example, "cat", "dog", "mouse", "cottage") and the class assigned to the image is "animals", traditional classification metrics may not show how well the model has actually learned. This limitation poses challenges for the~manual analysis, making it difficult to~discover potential issues during model training. Our proposed method shows that the XAI Ensembler can have the capability to explain what the underlying model has learned from the~explanations and~highlight disparities from the~expected outcomes. Through the analysis of these distinctions, it allows for a~precise understanding of~rare or peculiar images, different from the average content of most of the other images, and~the~reasons behind their challenges.

Furthermore, to~address the~performance disparities between test and~training data, we introduce the~\textit{data-oriented diverseness (div)} metric. This metric serves as an~indicator of~a~dataset's quality, determining whether a~sufficiently diverse set of~images is present to~train a~model with a~robust representation. Denoted as \textit{div(m)}, it quantifies the~similarity between validation and~training performance metrics. Specifically, for IoU, the~formula is expressed as:

\begin{equation}
div(iou) = 1 - ens_{train}(iou) + ens_{valid}(iou).
\end{equation}

Low values of~\textit{div(iou)} suggest that there are substantial differences between the~model's ensembler performance on training and~testing datasets. This can be an~indication of~potential issues in~the dataset, such as insufficient image diversity, leading to~challenges in~training a~model that generalizes well across various scenarios.

\textbf{Explanation-oriented exhaustiveness metric}
\nopagebreak

The~\textit{explanation-oriented exhaustiveness (exh)} metric helps to assess the~completeness of~an~explanation relative to~an entire input image, measuring the~difficulty of~extracting the~object of~interest in~explanations compared to~segmenting it in~the original image. Specifically, for the~Intersection over Union (IoU) metric, the~formula is given by:

\begin{equation}
{exh}(iou) = \frac{ens(iou)}{IoU (I_i, \mathcal{M})}.
\end{equation}

This metric provides insights into how well the~generated explanations capture the~salient features of~an object within the~broader context of~the segmenting original image. A~higher \textit{exh} value indicates a~more exhaustive and~reliable explanation, emphasizing a~model's ability to~highlight important elements.

\begin{figure}
    \centering   
    \begin{minipage}{\linewidth}
      \centering
      \includegraphics[align=c, width=0.325\linewidth]{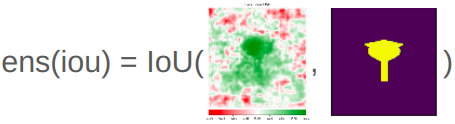}
    \includegraphics[align=c,width=0.325\linewidth]{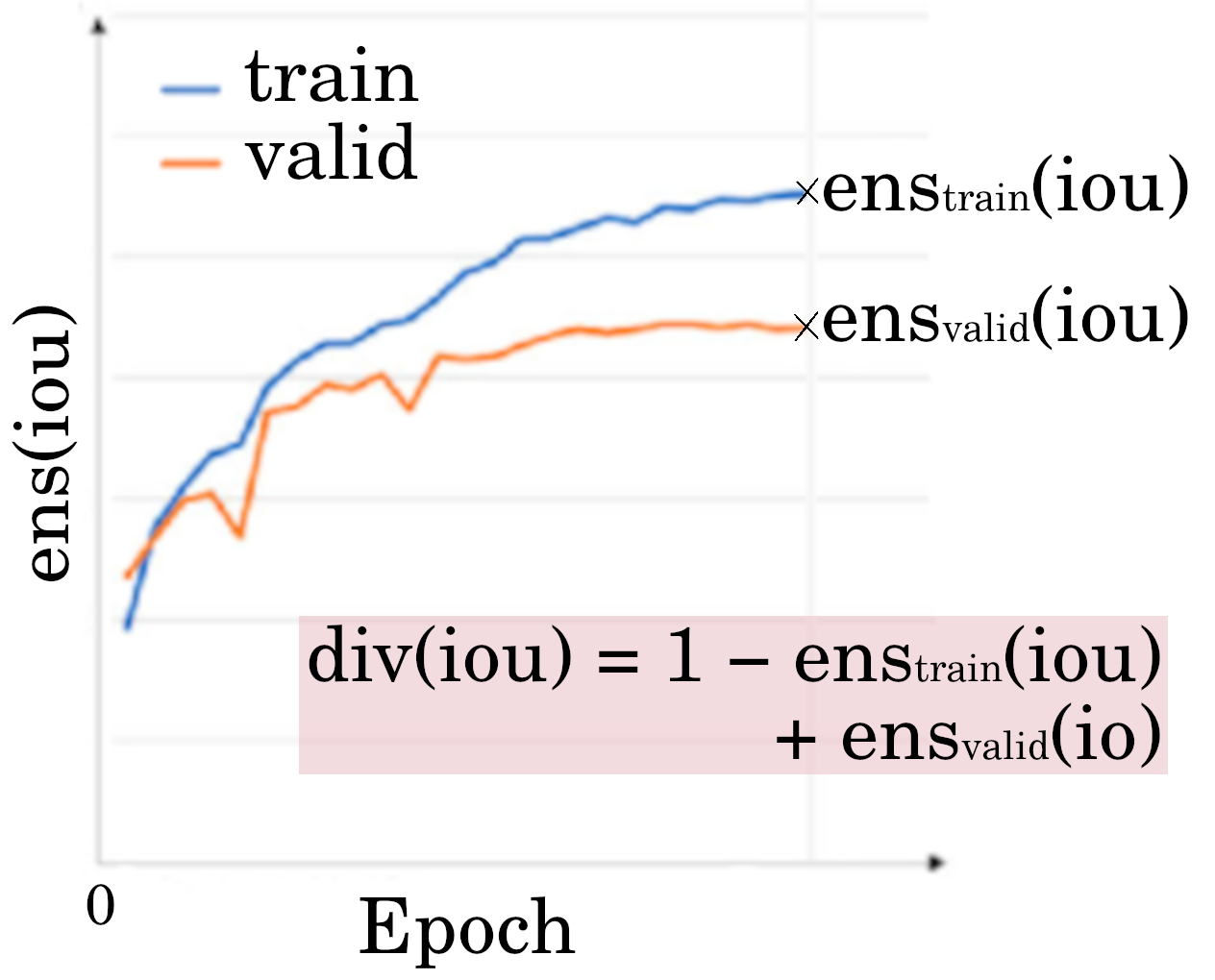}
    \includegraphics[align=c,width=0.325\linewidth]{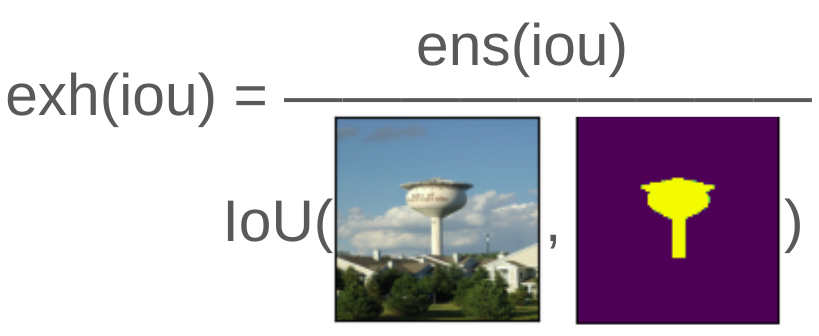}
      
    \end{minipage}

    \caption{A simplified illustration of introduced metrics: ensembling performance, diverseness and exhaustiveness.}
    \label{fig:metrics_summary}
\end{figure}

For each of the above metrics, the visuals in Figure \ref{fig:metrics_summary} highlight the actual problem and what they actually are aimed to measured. 

\subsection{Representation evaluation}

High accuracy on datasets like ImageNet may not be enough for thorough model evaluation~\cite{Beyer2020AreWD}. The~experiment focuses on adding an interpretability layer to enable a model's explanation, allowing a~deeper understanding of~the~learned representations beyond accuracy metrics. This aligns with the~perspective that assessing model performance requires considering not only predictive accuracy but also interpretability and~the~meaningfulness of~the~learned representations.

\begin{figure}[h]
    \begin{subfigure}[t]{0.25\textwidth}
        \includegraphics[height=4cm,valign=t]{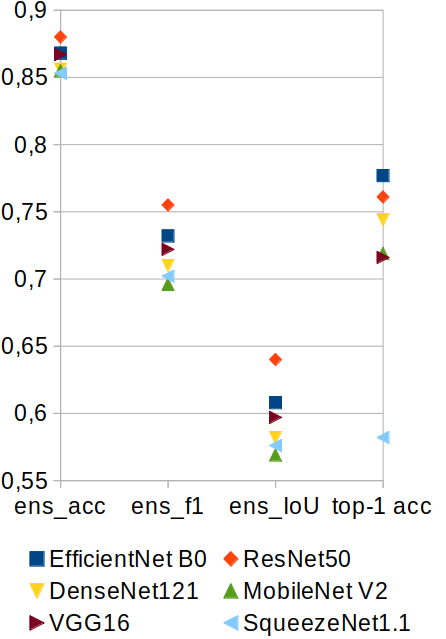}\vspace{1mm}
        \caption{Ensembling performance metric for different models.}
        \label{fig:representation}
    \end{subfigure}
    \hfill
    \begin{subfigure}[t]{0.3\textwidth}
        \hspace{-0.8cm}
        \includegraphics[height=4.1cm,valign=t]{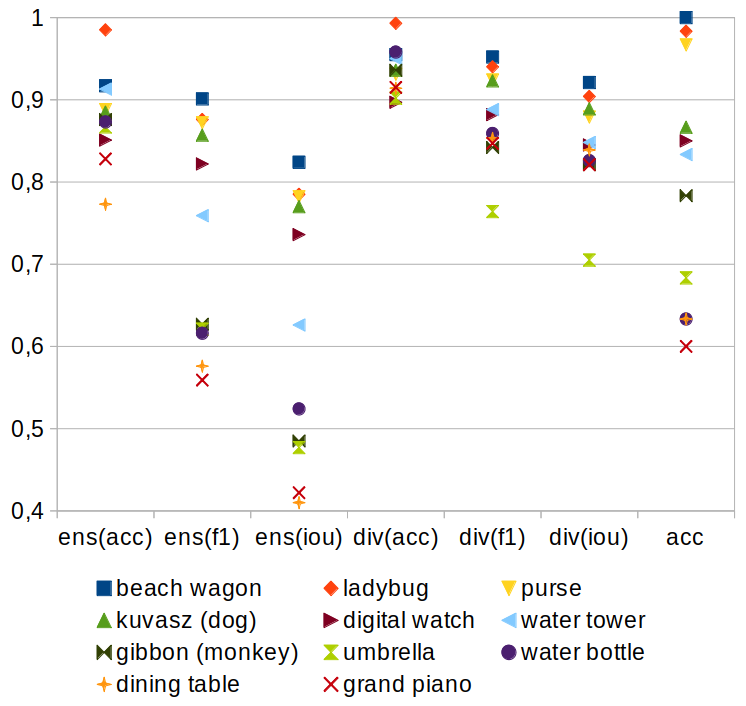}
        \caption{Ensembling performance, and~diverseness metric for different classes.}
        \label{fig:class}
    \end{subfigure}
    \hfill
    \begin{subfigure}[t]{0.42\textwidth}
        \includegraphics[height=3.6cm,valign=t]{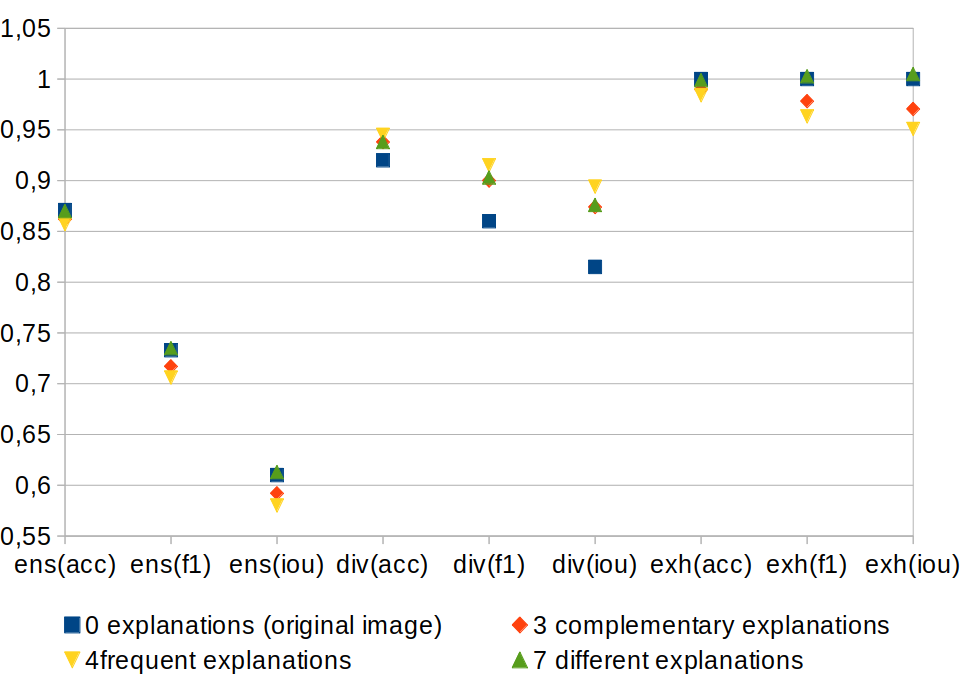} \vspace{4.5mm}
        \caption{Ensembling performance, diverseness, and~exhaustiveness metric for different sets of~explanations.}
        \label{fig:xai}
    \end{subfigure}
    \caption{Averaged results of~introduced metrics on the~testset.}
\end{figure}

This experiment aims to~determine which classification model from SqueezeNet1.1 MobileNet v2, VGG16, DenseNet121, ResNet50, and~EfficientNet B0 learned the~best representation. To achieve this, we trained XAI ensembling models using explanations generated by mentioned models. We decided to~average the~results for eleven classes and~three different sets of~explanations, listed in~legends to~Figures \ref{fig:class} and~\ref{fig:xai}, respectively. The~results are shown in~Figure~\ref{fig:representation}. Moreover, in~Figure~\ref{fig:representation} a~performance comparison of~various architectures on ImageNet dataset is presented, illustrating their top-1 accuracy.

From~Figure~\ref{fig:representation}, several key observations can be made. Namely, the~ensembling accuracy metric is the~least differentiating. This is in~line with assumptions, because segmentation models are evaluated at the~pixel level and~often there is a~significant class imbalance between the~object area and~the~background. ResNet50 outperforms all other models in~all ensembling performance metrics. However, in~classification metrics, it has lower accuracy than EfficientNet B0, which took second place in~ensembling performance metrics. The~order of~models according to~ensembling f1 and~IoU metrics is the~same. This is because both metrics are similar to~each other, but it is worth noting that they are very suitable for evaluating the~segmentation task. VGG proved better at learning representation than the~very deep DenseNet121 model and~two small models: MobileNet and~SqueezeNet. What may seem surprising is the~result of~SqueezeNet, which according to~ensembling performance metrics, positions itself higher than MobileNet, which contradicts the~accuracy metric, of~which it is more than 10\% lower.

To sum up, our experiment highlights the~strong performance of~ResNet50, a~trend consistent with its frequent use as a~backbone in~various computer vision tasks, including semantic segmentation (as seen in~Mask R-CNN~\cite{He_2017_ICCV}) and~object detection (as seen in~Faster R-CNN~\cite{NIPS2015_14bfa6bb}). Researchers and~practitioners often leverage its capabilities as a~feature extractor~\cite{obdet}, utilizing pre-trained ResNet weights to~boost the~performance of~their models on specific tasks.

\subsection{Dataset evaluation}

In~this experiment, our objective was to~investigate whether the~difficulty of~an~image class from the~ImageNet dataset could be evaluated through explanation ensembling. We trained CNN-based explanation ensembling on eleven various image classes using 5-fold cross-validation, explanations generated by ResNet50, and using three sets of explanation methods. Cross-validation is important especially for dataset evaluation, as it reduces the impact of splitting the dataset into training and test data. The~averaged across cross-validations performance metrics are shown in~Figure \ref{fig:class}.

\begin{figure}[h]
    \centering
    \includegraphics[width=\linewidth]{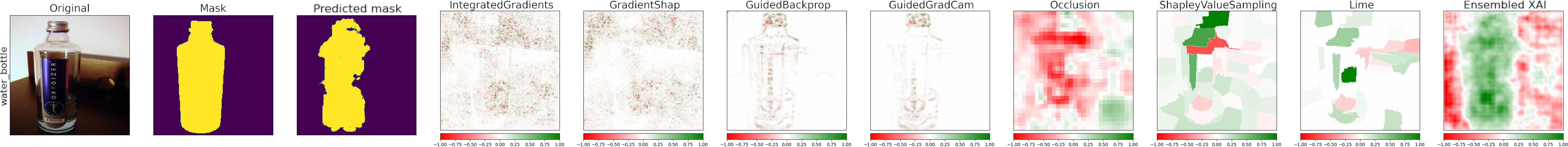}
    \includegraphics[width=\linewidth]{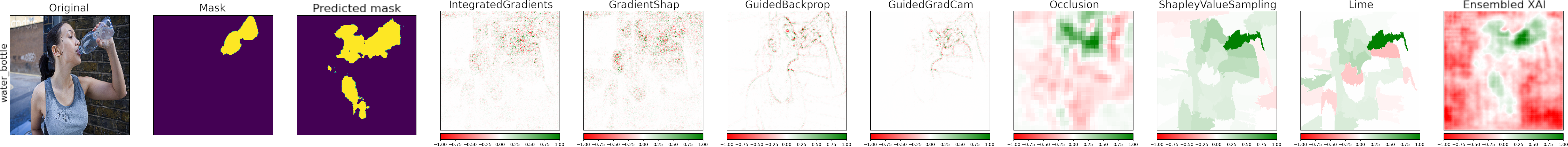}
    \caption{Two examples from the~ImageNet dataset belonging to~the "water bottle" class. The~first example is commonly found in~the dataset, while the~second example is challenging and~infrequently encountered. Each example consists of~an~image, a~mask, a~predicted mask, and~individual component explanations. The~final component is the~ensembled explanation.}
    \label{fig:bottle}
\end{figure}

By analyzing the~performance metrics, we can gain insights into the~difficulty of~each image class. Firstly, identifying classes with the~lowest \textit{ens(f1)} and~\textit{ens(IoU)} highlights challenging classes, such as "dining table" and~"grand piano". Addressing the~difficulties in~recognizing these classes may be crucial for applications like training models for object recognition in~autonomous vacuum cleaners. Particularly, in~some cases, it is important to~ensure fairness by preventing subjects such as "monkeys" from being misidentified as "humans". Secondly, it is important to examine the~cases where the~greatest difference in~the \textit{ens(f1)} and \textit{ens(IoU)} between the~training and~testing sets occurs. Taking the~"water bottle" as an~example, presented in~Figure \ref{fig:bottle}, understanding such disparities is essential, especially if the~context of~the images in~the dataset significantly differs between themselves, as observed in~the varying poses and~contexts of~"water bottles".

The~last column in~Figure \ref{fig:class} presents the~accuracy of~the~classification task. The~correlation can be seen between the~individual class score in~the introduced metrics (ensembling performance and~diverseness) and~the~accuracy score. From~the~obtained results, it can be seen that classes that get high scores in~the~classification usually get high scores in~our metrics. Our metrics have the~advantage over the~simple classification metrics, because their results can be easily visualised as masks or heatmaps, which help to~find the~reason for a~potential misclassification error.

By leveraging explanation ensembling, we can gain a~holistic understanding of~the~difficulty level associated with different image classes. This information can be valuable for various applications, such as prioritizing resources for challenging image classes, improving model performance on difficult classes, or tailoring training strategies to~address class-specific difficulties.

\subsection{Explanations evaluation}

In~this section, we present the~experimental results of~our study, which aimed to~answer the~research question of~how well the~explanations succeeded in~extracting relevant features. We also determine the~influence of~each input explanation on the~final result.

The experimental setting is the same as in the previous subsection. To understand the~significance of~the number of~XAI methods incorporated in~the CNN-based explanation ensembling, we visualised the~results in~Figure~\ref{fig:xai}. The~measure \textit{exh(iou)} determines how much of~the important features are left in~the picture. Our analysis revealed that the~results of~segmenting an~original image and~applying XAI ensembling on seven different explanations obtained almost the~same results. 

These findings indicate that, based on the~selected performance metric, seven XAI methods have the~potential to~deliver the~same predictive performance as the~original images. The~original image contains many more features than its explanations. Demonstrating that explanations can serve as substitutes for original images, this approach holds promise for facilitating reduction of~redundant information or the~anonymization process of images. Anonymization involves transforming data into a~format that does not disclose individuals' identities. In~this context, one possible application could be modifying existing images~\cite{Wang2022}, or alternatively, employing synthetic data generation as an~alternative method for data anonymization.

Several studies report that no single explanability method performs best in~all metrics evaluating the~quality of~an~explanation~\cite{Zhou_Booth_Ribeiro_Shah_2022, Komorowski_2023, pmlr-v162-kim22h}. However, it is hard to~clearly state how explanability methods should be compared among themselves. It is possible to~choose a~single metric, but according to~each metric, the~result may be different. Faced with this, comes the~problem of~evaluating the~validity of~metrics for evaluating explanations among themselves. 

In~our solution, we do not have such a~problem. Our explanation ensembling solution allows, after training the~XAI Ensembler, to~disable, in other words set the~attribution values to~zero, individual input explanations. Disabling one component explanation can measure its validity in~the~aggregated explanation. For~this reason, we check the~impact of~different explanability methods on the~ensembling result. 

In~Table \ref{tab:xai_investigation}, the~"difference" columns show the~average difference in~\textit{ens(acc)}, \textit{ens(f1)} and~\textit{ens(IoU)} between the~model with all XAI methods and~the~model with the~specified XAI method removed. The~"quotient" columns show the~average ratio of~the performance of~the model with all the~XAI methods to~the performance of~the model with the~specified XAI method removed. 

As Table \ref{tab:xai_investigation} shows, the~GuidedBackprop explanation contributes the~most. What is worth noting is that the~explanations are complementary. After removing one of~them, there is no significant drop in~the quality of~the generation of~the corresponding mask (the difference in~the resulting mask when giving all explanations in~the input and~when giving all of~them without one explanation). It is important to~note that these results are based on a~ResNet50 model and~may not generalize to~other models or datasets.

\begin{table}[h]
    \centering
    \begin{tabular}{l|rrr|rrr}
    \multirow{2}{*}{\textbf{Removed XAI}} & \multicolumn{3}{c|}{\textbf{Difference}} & \multicolumn{3}{c}{\textbf{Quotient}}  \\
    & \textbf{ens(acc)} & \textbf{ens(f1)} & \textbf{ens(IoU)} & \textbf{ens(acc)} & \textbf{ens(f1)} & \textbf{ens(IoU)} \\
    \midrule
    GuidedBackprop & 0,0478 & 0,0770 & 0,0964 & 0,9465 & 0,8937 & 0,8501 \\
    GuidedGradCam & 0,0208 & 0,0413 & 0,0482 & 0,9763 & 0,9389 & 0,9156 \\
    ShapleyValueSampling & 0,0086 & 0,0113 & 0,0135 & 0,9899 & 0,9667 & 0,9586 \\
    Occlusion & 0,0072 & 0,0110 & 0,0116 & 0,9917 & 0,9666 & 0,9600 \\
    GradientShap & 0,0037 & 0,0037 & 0,0035 & 0,9963 & 0,9874 & 0,9853 \\
    Lime & 0,0028 & 0,0004 & 0,0003 & 0,9972 & 0,9906 & 0,9901 \\
    IntegratedGradients & 0,0025 & -0,0005 & -0,0012 & 0,9974 & 0,9911 & 0,9912 \\
    \bottomrule
    \end{tabular}
    \caption{Impact of~removing XAI methods on ensemble model performance. The~"difference" columns show the~average decrease in~performance metrics, while the~"quotient" columns express the~remaining performance as a~ratio compared to~the model with all XAI methods.}
    \label{tab:xai_investigation}
\end{table}


\section{Conclusions and~future works}

The~novel aspect of~our CNN-based explanation ensembling lies in~using explanations to~evaluate the~learned representation. Our approach provides a~unique way to~assess the~difficulty of~each class and~the~quality of~trained models. It aids in~the selection of~a~pretrained backbone for tasks like detection, providing quantitative evaluation and~dataset analysis. Noteworthy is the possibility of ensembling explanations from trained vision transformers, as their explanations are also commonly presented as heatmaps \cite{stassin2023explainability}.

We gain insights into where the~model excels and~where it struggles by analyzing the~performance metrics of~the~generated explanations for many class. Through the~study on the~elements that should or should not be present in~the ensembled explanation, our method helps to~predict the~possibility of~bias, which is particularly crucial for sensitive attributes, and~suggests strategies to~address imbalances in~the dataset. This valuable information can guide further improvements in~dataset preparation and~model training. Our approach finds unnecessary information in~data, which may help in~designing algorithms that enhance anonymity.

Morever, our CNN-based explanation ensembling presents a~novel approach to~address the~challenge of~generating comprehensive explanations, in~which we do not have to~worry about the~importance of~each of~the aggregated explanation methods. Through the~use of~carefully selected evaluation metrics from the~Quantus library, we demonstrated the~method's superior performance in~terms of~Localisation and~Faithfulness, compared to~individual XAI methods.

While we have highlighted the~key strengths of~our approach, it is essential to~acknowledge its limitations. The~performance of~our method is highly dependent on the~quality and~diversity of~the training data, as limited or biased training data can result in~suboptimal explanations and~potentially undermine a~model's reliability. Furthermore, ensembling explanations using CNNs can be computationally intensive, which may limit its applicability in~resource-constrained environments. Interpretability remains an~ongoing challenge, as even the~best explanations might not fully capture the~complex inferential processes of~deep learning models. It is crucial to~recognize these limitations to~ensure the~responsible and~informed use of~the CNN-based explanation ensembling.

In~light of~these limitations, future research can explore several directions. First of~all, we see a~growing area of~universal, robust feature extractors, such as DINOv2~\cite{oquab2023dinov2}. We believe that the~CNN-based explanation ensembling might be available as pretrained backbone and work on many different classes. Additionally, an idea of the~trainable XAI Ensembler can be generalised to other modalities. The~code for our CNN-based explanation ensembling is available at \\ \url{https://github.com/Hryniewska/CNNbasedXAIensembling}.


\section*{Acknowledgment}
Thank you to~Szymon Bobek for his feedback, which helped to~improve the~paper. We are also grateful to anonymous reviewers for their insightful suggestions. The work was financially supported by the~NCBiR grant INFOSTRATEG-I/0022/2021-00, and carried out with the~support of~the Laboratory of~Bioinformatics and Computational Genomics and the~High Performance Computing Center of~the Faculty of~Mathematics and Information Science, Warsaw University of~Technology.





\small
\bibliographystyle{unsrtnat}
\bibliography{references}  

\end{document}